\documentclass{article}

\usepackage[nonatbib, preprint]{main}

\usepackage[utf8]{inputenc} 
\usepackage[T1]{fontenc}    
\usepackage[hidelinks]{hyperref}       
\usepackage{url, caption, subcaption}   
\usepackage{booktabs, appendix}       
\usepackage{amsfonts}       
\usepackage{nicefrac}       
\usepackage{microtype, xcolor, soul}      
\usepackage[ruled]{algorithm2e}
\usepackage{amsmath,amssymb,latexsym,amsthm}
\usepackage{graphicx,psfrag,epsf, csquotes,float}
\usepackage{wrapfig}

\definecolor{lightblue}{rgb}{0.188, 0.478, 0.858}
\definecolor{brown}{rgb}{0.76,0.5,0.18}
\newcommand{\refC}[1]{\textcolor{red}{\ref{#1}}}
\newcommand{\Ccite}[1]{\textcolor{lightblue}{\cite{#1}}}
\newcommand{\algo}[1]{\textcolor{brown}{\selectfont\ttfamily{{#1}}}}

\newtheorem{lemma}{Lemma}

\title{Learn Dynamic-Aware State Embedding for Transfer Learning}

\author{%
     Kaige Yang\\
    University College London\\
    \texttt{Kaige.yang.11@ucl.ac.uk}\\
}

\begin{document}

\maketitle

\begin{abstract}
    Transfer reinforcement learning aims to improve the sample efficiency of solving unseen new tasks by leveraging experiences obtained from previous tasks. We consider the setting where all tasks (MDPs) share the same environment dynamic except reward function. In this setting, the MDP dynamic is a good knowledge to transfer, which can be inferred by uniformly random policy. However, trajectories generated by uniform random policy are not useful for policy improvement, which impairs the sample efficiency severely. Instead, we observe that the binary MDP dynamic can be inferred from trajectories of any policy which avoids the need of uniform random policy. As the binary MDP dynamic contains the state structure shared over all tasks we believe it is suitable to transfer. Built on this observation, we introduce a method to infer the binary MDP dynamic on-line  and at the same time utilize it to guide state embedding learning, which is then transferred to new tasks. We keep state embedding learning and policy learning separately. As a result, the learned state embedding is task and policy agnostic which makes it ideal for transfer learning. 
    In addition, to facilitate the exploration over the state space, we propose a novel intrinsic reward based on the inferred binary MDP dynamic. Our method can be used out-of-box in combination with model-free RL algorithms. We show two instances on the basis of \algo{DQN} and \algo{A2C}. Empirical results of intensive experiments show the advantage of our proposed method in various transfer learning tasks.
\end{abstract}

\section{Introduction}
Deep reinforcement learning has gone through a rapid development in recent years and show near-human or super-human performance in many domains like games \Ccite{silver2017mastering}, protein-folding \Ccite{senior2020improved} and robotic \Ccite{levine2018learning}. Despite countless success, there still remains many challenges. One is the limited flexibility in solving different tasks. Human excel at adjusting behaviour in new tasks when the goal is changing. This ability relies on the efficient transfer of knowledge accumulated in one task  to others. 

Transfer learning in RL aims to leverage the acquired knowledge from seen tasks to efficiently solve unseen tasks. In order to generalize to unseen tasks, an agent needs to abstract the general aspect of tasks. In this work, we are interested in a particular instance of transfer learning setting where all tasks (MDPs) are the same except the reward functions. In this setting, one of the general aspect of tasks is the shared dynamic of MDPs. However, the MDP dynamic is typically unknown or requires a large amount of data to learn and prone to error. For example,  \Ccite{mahadevan2007proto} and \Ccite{madjiheurem2019representation} proposed to learn the MDP dynamic via uniform random policy. To obtain an accurate estimation, it requires a large amount of interactions between agent and the environment, which is infeasible in case of large state space. Moreover, the data generated by uniform random policy can not be utilized for policy improvement. This impairs the sample efficiency severely. 

Instead, we observe that the binary MDP dynamic, unlike MDP dynamic, can be inferred from trajectories generated by any policy. This avoids the need of uniform random policy and allows the agent to infer the dynamic as the same time of improving policy. The transition matrix of the binary MDP dynamic is the binary version of the transition matrix of MDP dynamic. To estimate it, a significant less amount of data is required.  As the binary MDP dynamic contains the information of state structure which is shared by all tasks, we believe such knowledge is suitable to transfer.

Built on the observation, we design an agent that can infer the binary MDP dynamic online and leverage it to guide state embedding, which is then transferred to new tasks. Specially, we propose a method to learn the binary MDP dynamic gradually from trajectories 
generated by policy iteration procedure. At the same time, we utilize it to guide state embedding learning in a way that close states have similar embedding. This is based on the intuition that in general consecutive states have similar state values. We keep the state embedding learning and policy learning separately. This ensure the policy-dependency of state embedding minimal which improves its ability for transfer learning. 

In addition, to have an accurate estimation of the binary MDP dynamic, it is crucial that the agent 
visit the whole state space. However, exploration in RL is a long-standing challenge. To encourage the agent to visit more states, we propose a novel intrinsic reward based on the inferred binary MDP dynamic. Namely, an action is rewarded if the next state has been visited limited times  and has few neighbors. Empirical results show that the intrinsic reward encourages the agent to explore significant larger state space than $\epsilon$-greedy, which in turn speeds up the dynamic learning, state embedding learning and policy learning. 

Our methods can be applied generally to various model-free RL algorithms and requires minimal changes of RL framework. The main contributions of this work can be listed as follows:
\begin{itemize}
    \item We introduce a method to infer the binary MDP dynamic gradually based on policy iteration trajectories. 
    \item We propose a loss function to learn state embedding in line with the inferred dynamic.
    \item We propose a novel intrinsic reward to encourage exploration.
    \item Through intensive experiments, we show the advantage of our methods over baselines on various transfer learning tasks.
\end{itemize}
The rest of this paper is organized as follows. Section 2 presents an overview of the related work. In Section 3, the setting of reinforcement learning and transfer learning are described. Section 4 introduces the binary MDP dynamic and the proposed method to infer it. Section 5 presents the proposed state embedding learning method. Section 6 describes the proposed intrinsic reward. In section 7, two proposed algorithms are presented. Experiment results are reported in Section 8. Section 9 includes the conclusion and future research directions.

\section{Related Work}
A number of works have focused on learning state embedding in line with the underlying shared MDP dynamic across tasks to facilitate the transfer learning ability. These works can be roughly categorized into two lines of research: 1), first infer the transition matrix of the MDP dynamic via uniform random policy and then learn state embedding through graph-based node embedding techniques; 2), learn successor representation (SR) or successor feature (SF) of good policies in seen tasks and transfer the learned SR/SF to unseen tasks. Here, we intensively discuss these research.

\Ccite{mahadevan2007proto} proposed to learn the transition matrix through uniform random policy. The transition matrix is viewed as the adjacency matrix of state graph where nodes represent states and edge weights represent the transition probability. Then, the eigenvectors of Laplacian, named \textit{proto-value-function (PVF)}, are used as state embedding. As PVF encodes the spectral structure of the state graph, make it suitable for transfer learning. The eigendecomposition of graph Lapalcian is infeasible in case of large state space which limits it applicability in practice. \Ccite{madjiheurem2019representation} proposed to learn transition matrix as \Ccite{mahadevan2007proto}. But, the state embedding is learned through \algo{node2vec} \Ccite{grover2016node2vec} algorithm. Beside \algo{node2vec}, many other node-embedding techniques can be utilized like \algo{DeepWalk}\Ccite{perozzi2014deepwalk} and \algo{GraphSage}\Ccite{hamilton2017inductive} in \Ccite{waradpandegraph}. As discussed before, the transition matrix requires a large amount of data to estimate and prone to error. More importantly, the data generated by uniform random policy is not useful for policy improvement, which impairs the sample efficiency seriously. 

In contrast, our work aims to infer the binary transition matrix which requires much less data to estimate. Furthermore, it can be inferred from data generated by any policy. This allows the agent infer the binary transition matrix at the same time of improving policy. 

An alternative line of research is based on successor representation (SR) \Ccite{dayan1993improving}. In tabular case, SR decouples the state value into future state occupancy and rewards which makes it suitable for transfer learning in fixed MDP dynamic. \Ccite{barreto2017successor} generalized SR to function approximation case, named successor feature (SF). The state value is decomposition into representation of future state feature occupancy (SF) and rewards. SF is dependent on the MDP dynamic and behaviour policy and is task-independent, while rewards are purely task-dependent. Thus, when SF under a policy is learned, it is possible to quickly evaluate the state value of the same policy under new tasks. However, the policy-dependency of SR/SF limits their applicability in transfer learning for two reasons. First, as pointed in \Ccite{lehnert2017advantages} \Ccite{madarasz2019better} a good policy in one task may performs poorly in another task. It is possible that important states in new tasks are not desirable in previous task. Thus, SR/SF of such states are not well represented by good policy in old tasks. This makes it hard for the agent to find optimal routes in new tasks.
Second, SR/SF are excel at evaluating a given policy under different tasks. However, to solve a task, the agent follows generalized policy iteration (GPI) \Ccite{sutton2018reinforcement}, where the policy keeps improving rather than remains the same. It means the SR/SF of one particular policy is quickly out-of-date as the policy is changing. To overcome these limitations, several attempts have been made. \Ccite{madarasz2019better}
proposed successor maps. \Ccite{ma2020universal} proposed Universal Successor Feature (USF).

Broadly speaking, SR/SF is a state embedding encoding the MDP dynamic and policy jointly. To avoid negative transfer in unseen tasks, the policy-dependency of SR/SF need to be carefully dealt with. In contrast, our work learns state embedding based on binary MDP dynamic, which is policy agnostic. This ensures positive transfer to unseen tasks as long as under the same MDP dynamic.

As we also propose a novel intrinsic reward to encourage exploration, we briefly review related works. RL agents rely on reward provided by environment to update its value function or policy. 
Thus it is essential for an agent to explore the state space efficiently to come cross rewards. Intrinsic rewards have been proposed to encourage agent to visit novel states. \Ccite{pathak2017curiosity} proposed curiosity-based approach where a model is trained to predict the next state given the current state and action. The prediction error is used as the intrinsic reward. In such way, the agent is encouraged to visit states the model is uncertain (high prediction error).
\Ccite{bellemare2016unifying} \Ccite{badia2020never} proposed count-based approaches where the inverse of state visitation count is used as intrinsic reward to encourage agent to explore less visited states.
\Ccite{marino2018hierarchical} proposed to reward an action if the representation of the current state and next state are significant different.
\Ccite{tao2020novelty} defined the novelty of a state as the sparsity in the area around the state in latent state space.

Unlike existing works, our proposed intrinsic reward based on state-count and the number of neighbors of each state. The inferred binary MDP dynamic can be viewed as a graph of states. The neighbors of each state is defined as the one-step reachable states. An action is rewarded if the next state has been visited less times and has few neighbors. Intuitively, if a state has no or few neighbors, it is good indicator that the space around this state is not well explored. To the best of our knowledge, our work is the first one leverages the neighbors of states as intrinsic reward.

\section{Reinforcement Learning and Transfer Learning}
Consider a MDP denoted as tuple $\mathcal{M}=(\mathcal{S}, \mathcal{A}, \mathcal{P}, \mathcal{R}, \gamma)$ where $\mathcal{S}$ is the state space, $\mathcal{A}$ is the action space. The transition kernel $P(s'|s,a)$ gives the transition probability to state $s'$ given action $a$ is taken in state $s$. A policy $\pi(a|s)$ denotes the probability of action $a$ in state $s$. The state value $V(s)$ and state-action value $Q(s,a)$ under policy $\pi$ are defined as follows:
\begin{equation}
    V^\pi(s)=\mathbb{E}[\sum_{t=1}^\infty \gamma^t r(s_t)|s_0=s]
\end{equation}
\begin{equation}
    Q^\pi(s,a)=\mathbb{E}[\sum_{t=1}^\infty \gamma^t r(s_t)|s_0=s, a_0=a]
\end{equation}
The goal is to find the optimal policy $\pi^*$ whose V-value and Q-value function satisfy the \textit{Bellman Optimal Equation}:
\begin{equation}
    V^{\pi^*}(s)=r(s)+\gamma \mathbb{E}[V^{\pi^*}(s')]
\end{equation}
\begin{equation}
    Q^{\pi^*}(s,a)=r(s)+\gamma \max_{a'\in \mathcal{A}}\mathbb{E}[Q^{\pi^*}(s', a')]
\end{equation}

We are interested in a particular instance of the transfer learning problem, where the environment dynamic is fixed, but the reward function varies (e.g., goal location in maze navigation). Formally, 
define a set of MDP as $\{\mathcal{M}_1,\mathcal{M}_2,...,\mathcal{M}_K\}$ where each MDP is denotes as a tuple $\mathcal{M}_i=(\mathcal{S}_i, \mathcal{A}_i, \mathcal{P}_i, \mathcal{R}_i, \gamma_i)$. In our setting, all components of MDP are fixed, except for the reward function. i.e., $\mathcal{R}_i\neq \mathcal{R}_j, \forall (i,j)\in [K]$. We use function approximator $f(\cdot, \boldsymbol{\theta})$ to predict state value $V(s)$ where $\boldsymbol{\theta}$ is the function parameters. Denote $\phi(s)$ as state embedding, then $V(s)$ can be expressed as 
\begin{equation}
    V(s)=f(\phi(s), \boldsymbol{\theta})
\end{equation}
State value $V(s)$ is different under different policy and task. Denote $V_{\mathcal{M}_i}^\pi(s)$ as the state value under policy $\pi$ in task ${\mathcal{M}_i}$, which is 
\begin{equation}
    V_{\mathcal{M}_i}^\pi(s)=f(\phi(s), \boldsymbol{\theta}_{\mathcal{M}_i}^\pi)
\end{equation}
where $\boldsymbol{\theta}_{\mathcal{M}_i}^\pi$ is the function parameters specified to policy $\pi$ and task ${\mathcal{M}_i}$.

An ideal state embedding $\phi(s)$ should have properties: 1), state value $V(s)$ is a simple function of $\phi(s)$; 2), $\phi(s)$ is a good discriminator for the states. 
In this work, we use another function approximator $g(\cdot, \alpha)$ to learn state embedding where $\phi(s)=g(s, \alpha)$. We notice that consecutive states are in general have similar state values. Thus, it is intuitively to encode consecutive states with similar embedding. The information of state structure (which states are consecutive) is contains in the MDP dynamic. Define $\mathbf{M}$ as the transition matrix of a MDP where each entry of $\mathbf{M}$ is $M(s,s')=\sum_{a\in \mathcal{A}}\mathcal{P}(s'|s,a)$.  If $M(s,s')$ is known, we can enforce the similarity between state embedding in proportional to $M(s,s')$. e.g., the inner product $\phi(s)^T\phi(s') \propto M(s,s')$. 

In transfer learning, if we can learn state embedding $\phi(s)$ from one task $\mathcal{M}_i$, which is in line with the underlying dynamic $\mathbf{M}$, as all tasks share the same dynamic, such state embedding can be reused in solving unseen new tasks $\{\mathcal{M}_j\}, j\neq i$. 

However, the MDP dynamic $\mathbf{M}$ is typically unknown. Despite it can be estimated via uniform random policy, i.e, $M(s,s')\approx \sum_{a\in \mathcal{A}}\pi_0(a|s)\mathcal{P}(s'|s,a)$ where $\pi_0(a|s)=1/|\mathcal{A}|$, This requires a large amount of data and prone to error. More importantly, the data generated by uniform random policy is not helpful for policy improvement. This jeopardises the sample efficiency severally. To circumvent this issue, we observe that the binary transition matrix also contains the state structure and can be leveraged for state embedding learning. More importantly, the binary transition matrix can be inferred from data generated by any policy. This avoids the need of uniform random policy and allows the agent to infer the dynamic at the same of improving policy. We define the binary MDP transition matrix in next section.

\section{Binary MDP transition matrix}
We denote $\mathbf{M}$ as the MDP transition matrix where $M(s,s')=\sum_{a\in \mathcal{A}}\mathcal{P}(s'|s,a)$ is the transition probability from state $s$ to state $s'$. 
The binary MDP transition matrix $\bar{\mathbf{M}}$ is defined as 
\begin{equation}
 \bar{M}(s,s')=
 \begin{cases}
 1 & if \  M(s,s')>0\\
 0 & if \ M(s,s')=0
 \end{cases}
\end{equation}
Intuitively, $\bar{\mathbf{M}}$ is the skeleton of MDP dynamic $\mathbf{M}$, which contains the underlying state structure. 
We denote $\mathbf{M}^\pi$ as the policy transition matrix where $M^\pi(s,s')=\sum_{a\in\mathcal{A}}\mathcal{P}(s'|s,a)\pi(a|s)$. We point out a relationship between the binary MDP transition matrix $\bar{\mathbf{M}}$ and policy transition matrix $\mathbf{M}^\pi$.
\begin{lemma}
\label{lemma: lemma1}
Suppose $\bar{\mathbf{M}}$ and $\mathbf{M}^\pi$ are defined as above. The following holds for any policy $\pi$. 
\begin{equation}
\label{eq: eq1}
    \forall (s_i, s_j)\in \mathcal{S}, \ if \ \bar{M}(s_i,s_j)=0, \ then \ M^\pi(s_i, s_j)=0
\end{equation}
and 
\begin{equation}
\label{eq: eq2}
        \forall (s_i, s_j)\in \mathcal{S}, \ if \ M^\pi(s_i, s_j)>0, \ then, \ \bar{M}(s_i,s_j)=1
\end{equation}
\end{lemma}
In words, Eq.~\refC{eq: eq1} says that if there is no transition between state $s_i$ and $s_j$ in the underlyin MDP, there will be no such transition under any policy $\pi$. Eq.~\refC{eq: eq2} says that if under any policy $\pi$, there is a transition between $s_i$ and $s_j$, it must be true that $\bar{M}(s_i, s_j)=1$.

Lemma~\refC{lemma: lemma1} makes the foundation of this work, which means $\bar{\mathbf{M}}$ can be recovered from trajectories generated by any policy $\pi$ under the MDP. Formally,
$\bar{\mathbf{M}}$ can be recovered based on trajectories $\mathcal{T}=\{T_1,..., T_m\}$.
\begin{equation}
\label{eq: recover_m}
  \bar{M}(s_i, s_j) = 
  \begin{cases}
  1 & if \ \exists (s_i=s_{t}, s_j=s_{t+1}) \in \mathcal{T}\\
  0 & otherwise
  \end{cases}
\end{equation}
If the state space $\mathcal{S}$ is fully explored by $\mathcal{T}$, $\bar{\mathbf{M}}$ can be inferred accurately. However, in practice, fully explore the state space is infeasible when the state space is large. To facilitate the exploration, we propose a novel intrinsic reward in Section~\refC{section: ir}. Before that, in Section~\refC{section: se} we show how to utilize the inferred binary transition matrix for state embedding learning.

\section{State Embedding}
\label{section: se}
As the binary transition matrix represents the state structure, we propose a loss function to learn state embedding in align with the state structure. Suppose $|\mathcal{S}|=N$ is the number of states. Denote the state embedding matrix as $\mathbf{\Phi}\in \mathbb{R}^{N\times d}$ where $\boldsymbol{\phi}(s_i)\in \mathbb{R}^d$, the $i$-th row of $\mathbf{\Phi}$, is the embedding of state $i$. The embedding $\mathbf{\Phi}$ can be learned via neural network. The similarity between $\boldsymbol{\phi}(s_i)$ and $\boldsymbol{\phi}(s_j)$ can be measured by dot product $l(s_i, s_j)=\boldsymbol{\phi}(s_i)^T\boldsymbol{\phi}(s_j)$.
To enforce consecutive states have similar states and non-consecutive states have dissimilar embedding we convert the binary transition matrix $\bar{\mathbf{M}}$ into a new matrix $\mathbf{W}$ where each entry is defined as
\begin{equation}
    W(s_i, s_j)=
    \begin{cases}
    1 & if \ \bar{M}(s_i, s_j)=1\\
    -1 & if \ \bar{M}(s_i, s_j)=0
    \end{cases}
\end{equation}
Given $\mathbf{W}$, the state embedding can be learned through the loss function below
\begin{equation}
\label{eq: aux_loss}
    L_s=||\hat{\mathbf{W}}-\mathbf{W}||_2
\end{equation}
where $\hat{\mathbf{W}} = \mathbf{\Phi}\mathbf{\Phi}^T$ measures the similarity between states.

Moreover, to avoid two consecutive states have too similar embedding, we force embedding of all states to be at least $w$ apart as in \Ccite{tao2020novelty}, where $w$ is a hyper-parameter.
\begin{equation}
    L_{csc}=\max(||\boldsymbol{\phi}_i-\boldsymbol{\phi}_j||_2-w, 0)
\end{equation}
Note that these two loss function are not conflicted. As $L_s$ poses constraints on inner product distance, while $L_{csc}$ on $l_2$ distance.
The overall loss function for state embedding learning is $\mathcal{L}=L_s+L_{csc}$. 

Maintaining the matrix $\mathbf{W}$ is impractical in case of large state space. To circumvent this issue, we maintain a buffer $\mathcal{D}^t_s=\{s_0, ,..., s_t\}$ containing the unique states experienced so far. We also store the neighbors of each state $\mathcal{N}_t(s_i)$. 
At each train step, a batch of states $\mathcal{B}_s$ is sampled uniformly from $\mathcal{D}^t_s$. The associated binary transition matrix is constructed based on neighbors of each state in $\mathcal{B}_s$. 

\begin{figure}[ht]
    \centering
    \begin{subfigure}{0.6\textwidth}
        \includegraphics[width=\textwidth]{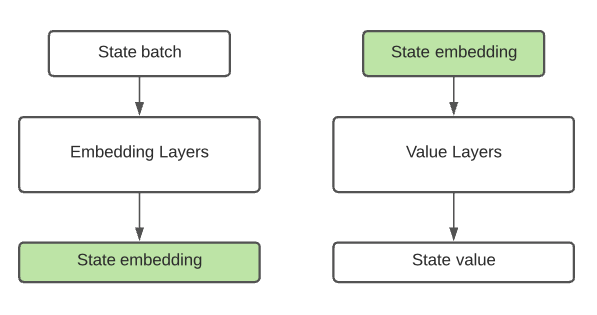}
    \end{subfigure}
\caption{Embedding module and value module}
\label{fig: module}
\end{figure}

It is worth to note that we keep the state embedding learning and state-value learning separately. This minimizes the policy-dependency of state embedding which facilitate its applicability for transfer learning. The neural network modules for embedding learning and state value learning are shown in Fig~\refC{fig: module}. The embedding module is trained by $\mathcal{L}=L_s+L_{csc}$ while the value module is trained by TD-error of state value.

\section{Intrinsic reward}
\label{section: ir}
To have an accurate estimation of the binary transition matrix, it is essential to explore the state space efficiently. 
To facilitate the exploration, we follow the works on intrinsic reward exploration, where the extrinsic reward is augmented with an intrinsic reward (exploration bonus). Specifically, we propose an intrinsic reward (IR) to encourage explore less visited states. The idea is that an action is rewarded if the next state has few neighbors and visited few times. Formally, denote $N_t(s)$ as the life-long visiting count which is the visiting count of state $s$ over the past episodes. Denote $d^e_t(s)$ as episodic-neighborhood namely the number of neighbors of state $s$ found within the current episode, which is reset as $1$ at the beginning of each episode. Formally, the intrinsic reward (IR) is defined as:
\begin{equation}
\label{eq: ir}
    \rho_t(s)=\frac{1}{\sqrt{N_t(s)d^e_t(s)}}
\end{equation}
Note that $\rho_t(s)$ is large when both $N_t(s)$ and $d^e_t(s)$ are small. A small $d_t^e(s)$ indicates the area around state $s$ is not well explored by this episode. We use it as an indicator for exploration. We also notice that some states may indeed have few neighbors in the underlying MDP (i.e., $d^e_t(s)$ remains small). In this case, $\rho_t(s)=1/\sqrt{d^e_t(s)}$ would encourage the agent to visit such states indefinitely. To avoid this issue, we make use of $N_t(s)$ the visiting count of state $s$ over all past episodes. This ensures that $\rho_t(s)$ approaches $0$ even $d_t^e(s)$ remains small.
The compound reward is defined as 
\begin{equation}
    r_t(s, a_t, s')=r^e_t+\beta \rho_t(s')
\end{equation}
where $r_t^e$ is the extrinsic reward provided by environment and $\rho_t$ is the intrinsic reward and $\beta$ is a scaling hyper-parameter.

\section{Algorithm}
The above described state embedding learning and intrinsic reward can be used in combination of any model-free RL algorithms. Here, we give an instance, named \algo{State2Emb+}, on the basis of \algo{DQN}. In the appendix, we provide another instance based on \algo{Actor-critic}.

\begin{algorithm}[ht]
\SetKwInOut{Input}{Input}\SetKwInOut{Output}{Output}
\Input{max episode number: $T_{max}$, hyper-parameter: $\beta$,  discount factor: $\gamma$, update frequency: $t_{freq}$}
\textbf{Initialization~~~}: agent policy $\pi$,  experience buffer $\mathcal{D}_{exp}$, state buffer $\mathcal{D}_s$, state neighbors
$\mathcal{N}(s)=\{ \}, \forall s\in \mathcal{S}$.
\BlankLine
\For{$t \leq T_{max}$}{
\begin{enumerate}
    \item Select action $a_t\sim \pi(s_t)$.
    \item Record transition $(s_{t}, a_t, s_{t+1}, r^e_t)$. 
    \item Update state visitation count $N(s_t)\gets N(s_t)+1$.
    \item Update state neighbors $\mathcal{N}_{s_t}\gets \mathcal{N}_{s_t}\cup \{s_{t+1}\}$ if $s_{t+1} \notin \mathcal{N}_{s_t}$.  
    \item Update $d_{s_t}=|\mathcal{N}(s_t)|$.
    \item Calculate intrinsic reward $\rho_t(s_{t+1})$ via Eq.~\refC{eq: ir}.
    \item Update state buffer $\mathcal{D}_s \gets \mathcal{D}_s \cup s_{t+1}$ if $s_{t+1}\notin \mathcal{D}_s$.
    \item Update eperience bufffer $\mathcal{D}_{exp}\gets \mathcal{D}_{exp}\cup \{(s_t, a_t, r^e_t, \rho_t(s_{t+1}), s_{t+1})\}$.
    \BlankLine
    \item \If{$t \ \ mod \ \ t_{freq} == 0$}{
    \begin{itemize}
       \item Sample experience batch $\mathcal{B}_{exp}=\{(s, a, r^e, \rho(s'), s')\}\in \mathcal{D}_{exp}$.
        \item Train value module with $\mathcal{B}_{exp}$.
        \item Sample state batch $\mathcal{B}_s=\{s_1,..., s_k\}\in \mathcal{D}_s$.
        \item Construct binary transition matrix $\mathbf{W}$ via $\{\mathcal{N}_{s_1},... \mathcal{N}_{s_k}\}$.
        \item Train value module with $\mathcal{B}_{s}$ via $\mathcal{L}=L_s+L_{csc}$.
    \end{itemize}
    }
    
    \end{enumerate}
}
\caption{\algo{State2Emb+}}
\label{algorithm: state2emb}
\end{algorithm}

\section{Experiments}
With experiments we aim to answer following questions: 1), does the proposed algorithm \algo{State2emb} perform better than baseline algorithms in transfers learning setting. 2), does the propose intrinsic reward (Eq.~\refC{eq: ir}) result in more efficient exploration?

\subsection{Transfer Learning}
To answer the first question, we test algorithms on transfer learning problems. Here, we describe the detail of experiment setting.

\begin{figure}[t]
    \centering
    \begin{subfigure}{0.32\textwidth}
        \includegraphics[width=\textwidth]{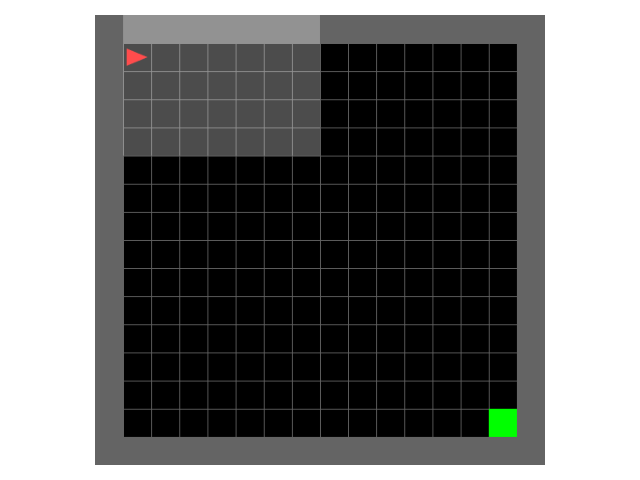}
        \subcaption[]{\textbf{ER}}
    \end{subfigure}
    \begin{subfigure}{0.32\textwidth}
        \includegraphics[width=\textwidth]{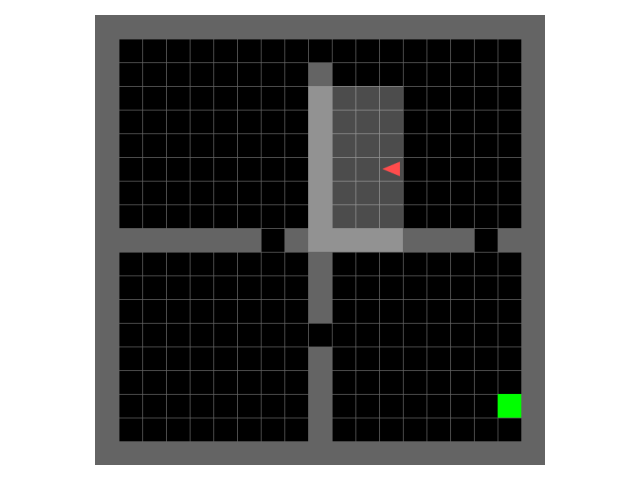}
        \subcaption[]{\textbf{FR}}
    \end{subfigure}
    \begin{subfigure}{0.32\textwidth}
        \includegraphics[width=\textwidth]{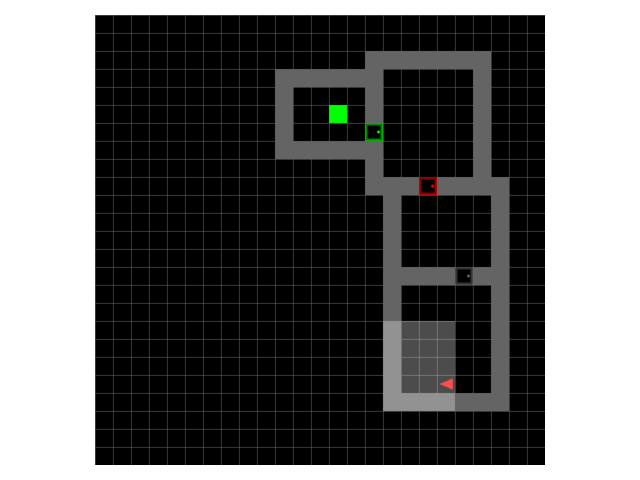}
        \subcaption[]{\textbf{MR}}
    \end{subfigure}
\caption{Navigation Tasks}
\label{fig: navg_tasks}
\end{figure}

\textbf{Environments:} We test algorithms on navigation tasks  \Ccite{gym_minigrid} with various difficulty. A easy task is empty-room (\textbf{ER)} shown in Fig~\refC{fig: navg_tasks}-a. A medium task is Four-room (\textbf{FR}) shown in Fig~\refC{fig: navg_tasks}-b. A hard task is Multi-room (\textbf{MR}) shown in Fig~\refC{fig: navg_tasks}-c.

\textbf{Experiment Setting:} We test algorithms on a set of tasks, where all task share the same MDP dynamic except the reward functions. This corresponds to different goal location in maze navigation tasks. The first task is treated as the source task from where algorithms learn state embedding. The rest are target tasks where the learned embedding remain fixed. In all tasks, the goal is to navigate to the goal as soon as possible (find the shortest path). Any episode terminates either the goal is arrived or the max step number $n_{max}$ is exhausted. No reward is given before arriving the goal. Upon arriving the goal the reward is defined as $1-0.9*(n_e/n_{max})$ where $n_e$ is the step number of current episode. Under such reward, long success episode is rewarded less than short success episode.

\textbf{Baseline Algorithms}: We compare the proposed algorithm \algo{State2Emb} and \algo{State2emb+} to baseline algorithms \algo{PVF}, \algo{state2vec}, \algo{SR}, \algo{DSF}, \algo{DQN} and \algo{Actor-Critic}. We do not include \algo{USF} and \algo{BSF} in baselines as these algorithms require the access of source target as well as target tasks when learn state embedding. While in our setting, target tasks are only accessible after source task. The state embedding is learned from the source task, which remains fixed in target tasks.

\begin{figure}[t]
    \centering
    \begin{subfigure}{0.45\textwidth}
        \includegraphics[width=\textwidth]{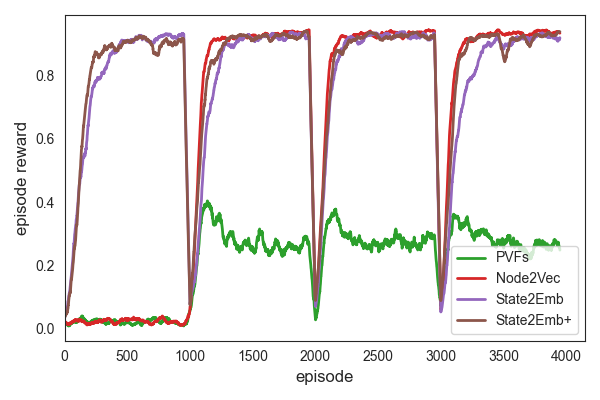}
        \subcaption[]{}
    \end{subfigure}
    \begin{subfigure}{0.45\textwidth}
        \includegraphics[width=\textwidth]{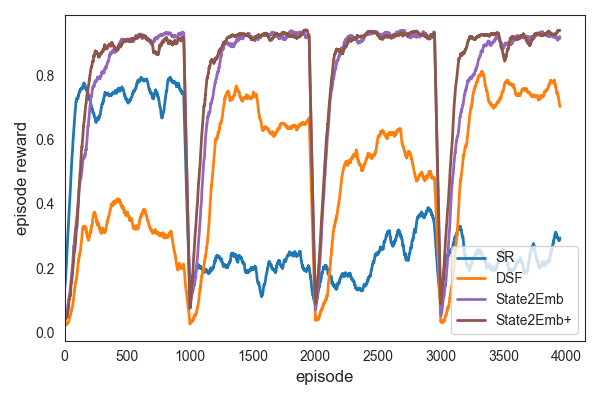}
        \subcaption[]{}
    \end{subfigure}
\caption{\textbf{ER}}
\label{fig: er}
\end{figure}

In Fig~\refC{fig: er}, we show the performance of algorithms in \textbf{ER} (results of \textbf{FR} and \algo{MR} will be shown in appendix). 
All algorithms are test on four consecutive tasks where the locations of start and goal are random selected as the beginning of each task. Each task lasts for $T=1000$ episodes. Fig~\refC{fig: er}-a compares the results of \algo{PVF}, \algo{Node2vec}, \algo{State2emb} and \algo{State2emb+}. The state embedding dimension is set as $d=10$ across all algorithms.

Several patterns are observed. First, during the source task (the first $1000$ episodes), the episodic reward of \algo{Node2vec} and \algo{PVF} remains low (almost zero). This is because that \algo{node2vec} and \algo{PVF} make use of uniform random policy to estimate the MDP dynamic. In contrast, \algo{State2emb} and \algo{State2emb} follow generalized policy iteration to find optimal policy for the source task and at the same time estimate the binary MDP dynamic. As is discussed before, binary MDP dynamic can be estimated following any policy which avoids the need of uniform random policy. Second, on the target tasks, \algo{state2emb} and \algo{state2emb+} have highly competitive performance with \algo{node2vec}. Note that the computational complexity of state embedding learning of \algo{node2vec} is much higher than that of \algo{state2emb}. Specifically, after learning the MDP dynamic, \algo{node2vec} unitizes \algo{skip-gram} to learn state embedding which involves simulating random walks and learn state embedding. In contrast, \algo{state2emb} learns state embedding based on sample batches via gradient descent at the same time of improving policy. Third, \algo{PVF} performs poorly with $d=10$. This is due to the fact that the state structure is represented by the full set of \textbf{proto-value-function} (PVFs) with dimension $d=N$ (the number of states). If we only use top-$10$ PVFs, a large part of information state-structure is missing. To have a better performance, a lager $d$ is required which impairs the computational efficiency.

Fig~\refC{fig: er}-b compares the performance of \algo{SR} and \algo{DSF}. It is clearly that \algo{state2emb} performs much better. The relative poor performance of \algo{SR} and \algo{DSF} is  due to the policy-dependency of SR and SF. As discussed before, at the end of the source task, SR and SF are learned based on the optimal policy of source task. However, the optimal policy of source task is not guaranteed to be good for target tasks. It is possible that important states in target tasks may not be well represented by the learned SR/SF which makes it hard for the agent to find optimal policy of target tasks. In contrast, \algo{state2emb} learns state embedding based on binary MDP dynamic which is policy diagnostic and guarantees the positive transfer to target tasks. 

Finally, \algo{state2emb+} performs better than \algo{state2emb} which is due to more efficient exploration as a result of the proposed instrinsic reward.

\subsection{Intrinsic Reward}
In this subsection, we examine the impact of the proposed intrinsic reward (IR) (Eq.~\refC{eq: ir}) on exploration and policy learning. We aim to answer the following questions: 1), does the IR encourage the agent to explore the state space more efficient than $\epsilon$-greedy? 2), does the IR keep pushing the exploration frontier forward? 3), does the IR speed up the convergence to optimal policy?

To answer the above questions, we test algorithms on \textbf{ER} task. Specifically, \algo{Q-Learning}, \algo{SR}, \algo{DQN} and \algo{AC} use $\epsilon$-greedy for exploration. Their counterparts \algo{Q-Learning+}, \algo{SR+}, \algo{DQN+} and \algo{AC+} employs $\epsilon$-greedy and the proposed IR. The $\epsilon$ probability is decayed following $\epsilon=0.9\times 0.95^t+0.1$, where $t$ is the episode index.
\begin{figure}[th]
    \centering
    \begin{subfigure}{0.45\textwidth}
        \includegraphics[width=\textwidth]{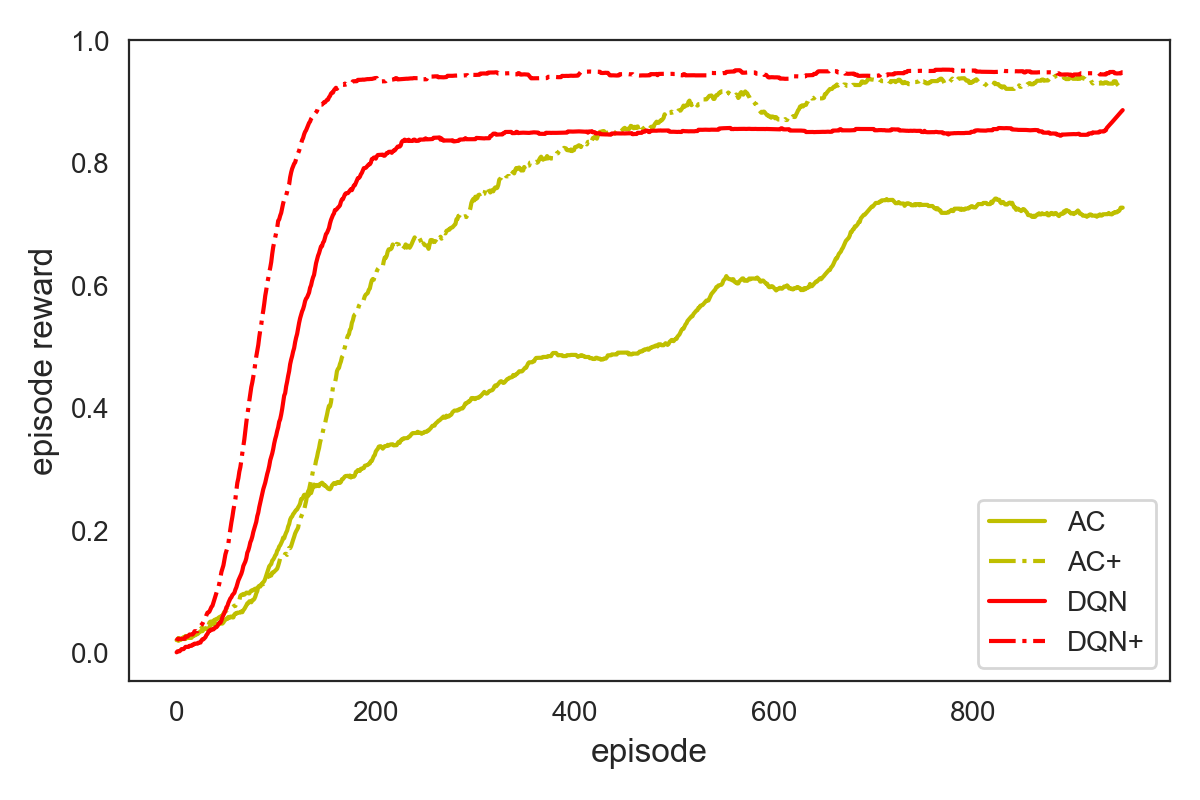}
    \end{subfigure}
    \begin{subfigure}{0.45\textwidth}
        \includegraphics[width=\textwidth]{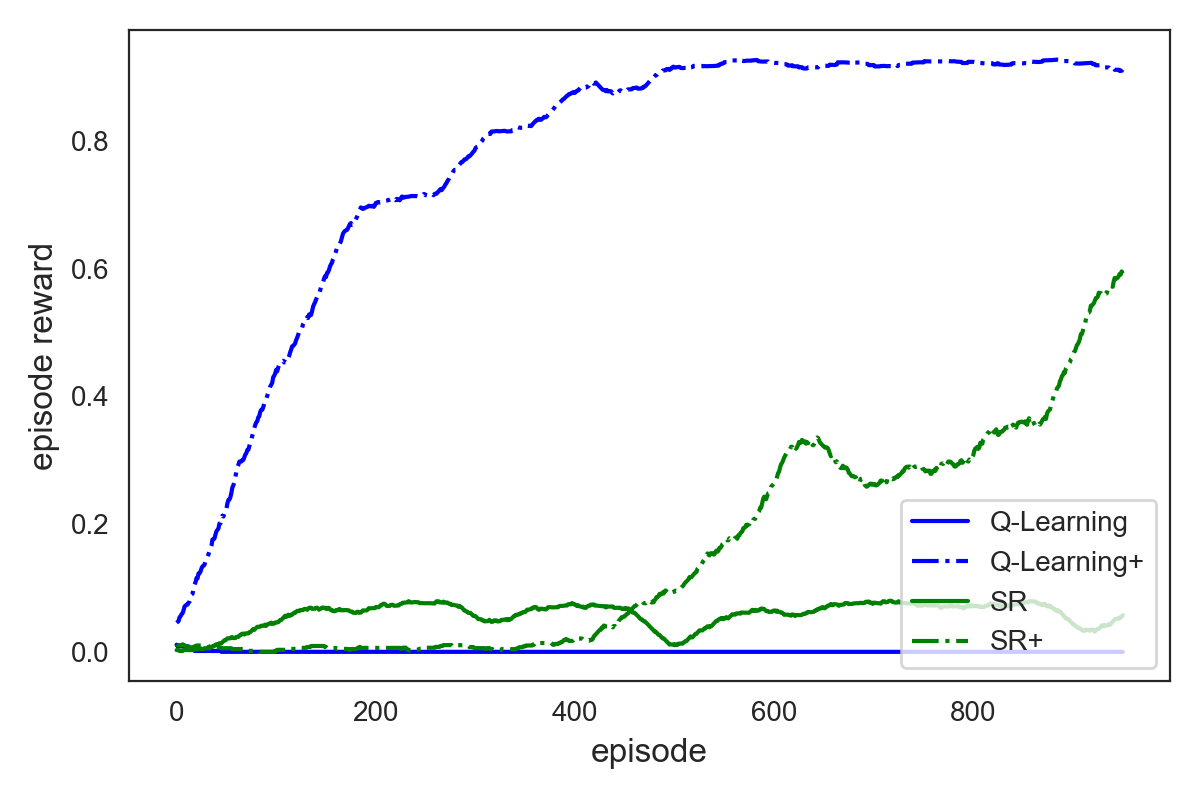}
    \end{subfigure}
\caption{Impact of IR on policy learning}
\label{fig: ir_imp}
\end{figure}

\begin{figure}[th]
    \centering
    \begin{subfigure}{0.32\textwidth}
        \includegraphics[width=\textwidth]{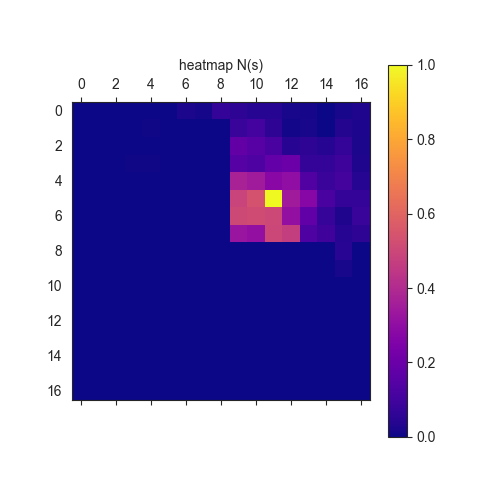}
        \subcaption[]{\algo{Q-Learning}}
    \end{subfigure}
    \begin{subfigure}{0.32\textwidth}
        \includegraphics[width=\textwidth]{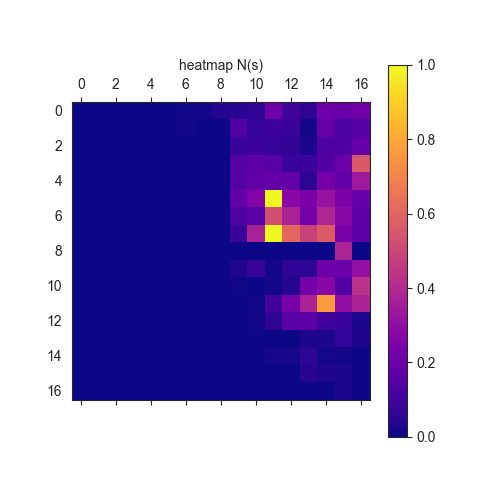}
        \subcaption[]{\algo{Q-Learning+}}
    \end{subfigure}
    \begin{subfigure}{0.32\textwidth}
        \includegraphics[width=\textwidth]{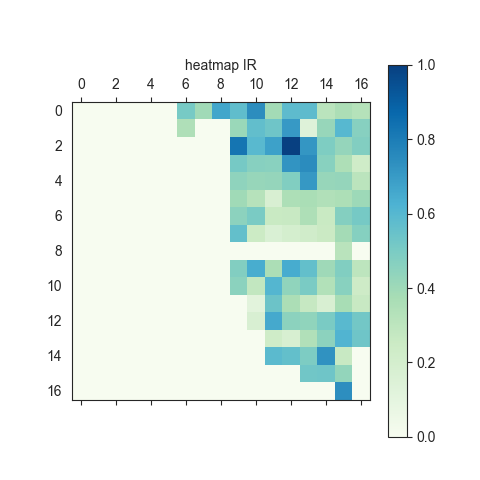}
        \subcaption[]{\algo{Q-Learning+}}
    \end{subfigure}
\caption{Heatmap of state visitation and IR}
\label{fig: heatmap}
\end{figure}

Fig~\refC{fig: ir_imp} compares the performance of algorithms with and without the proposed intrinsic reward. Apparently, all performances are enhanced by IR. The reason is simple more efficient exploration. To see this more clearly, Fig~\refC{fig: heatmap} shows the heatmap of state visitation count and IR. Specifically, Fig~\refC{fig: heatmap}-a  and Fig~\refC{fig: heatmap}-b show the heatmap of state visitation count, after $T=100$ episodes, of \algo{Q-Learning} and \algo{Q-Learning+}, respectively. Clearly, with the same number of episode, \algo{Q-learning+} explores more state space than \algo{Q-Learning}. The only difference between these two algorithms is the proposed intrinsic reward. Fig~\refC{fig: heatmap}-c shows the heatmap of IR. We can see that high value IR locates at less visited states which encourages the agent to keep explore new areas.

\section{Conclusion}
In this work, to solve transfer RL problme, we first point out the binary MDP dynamic can be inferred from any policy. As it contains the state structure information which is shared by all tasks, we proposed to learn state embedding in align with the inferred dynamic. Moreover, to encourage exploration we proposed a novel intrinsic reward based on the inferred binary dynamic. Through intensive experiments, we show the advantage of proposed algorithms in comparison with baselines. 
Several research directions can be explored. First, this work only applies to discrete state space case. It is an interesting direction to extend it to continuous state space. Second, the inferred binary MDP dynamic can be utilized for reward shaping or reward propagation. We believe this shall speeds up the policy convergence significantly.

\bibliographystyle{plain}
\bibliography{main}

\newpage
\section{Appendix}
\begin{algorithm}[ht]
\SetKwInOut{Input}{Input}\SetKwInOut{Output}{Output}
\Input{max episode number: $T_{max}$, hyper-parameter: $\beta$,  discount factor: $\gamma$, update frequency: $t_{freq}$}
\textbf{Initialization~~~}: agent policy $\pi$,  experience buffer $\mathcal{D}_{exp}$, state buffer $\mathcal{D}_s$, state neighbors
$\mathcal{N}(s)=\{ \}, \forall s\in \mathcal{S}$.
\BlankLine
\For{$t \leq T_{max}$}{
\begin{enumerate}
    \item Select action $a_t\sim \pi(s_t)$.
    \item Record transition $(s_{t}, a_t, s_{t+1}, r^e_t)$. 
    \item Update state visitation count $N(s_t)\gets N(s_t)+1$.
    \item Update state neighbors $\mathcal{N}_{s_t}\gets \mathcal{N}_{s_t}\cup \{s_{t+1}\}$ if $s_{t+1} \notin \mathcal{N}_{s_t}$.  
    \item Update $d_{s_t}=|\mathcal{N}(s_t)|$.
    \item Calculate intrinsic reward $\rho_t(s_{t+1})$ via Eq.~\refC{eq: ir}.
    \item Update state buffer $\mathcal{D}_s \gets \mathcal{D}_s \cup s_{t+1}$ if $s_{t+1}\notin \mathcal{D}_s$.
    \item Update eperience bufffer $\mathcal{D}_{exp}\gets \mathcal{D}_{exp}\cup \{(s_t, a_t, r^e_t, \rho_t(s_{t+1}), s_{t+1})\}$.
    \BlankLine
    \item \If{$t \ \ mod \ \ t_{freq} == 0$}{
    \begin{itemize}
       \item Sample experience batch $\mathcal{B}_{exp}=\{(s, a, r^e, \rho(s'), s')\}\in \mathcal{D}_{exp}$.
        \item Train value module with $\mathcal{B}_{exp}$.
        \item Train policy module with $\mathcal{B}_{exp}$
        \item Sample state batch $\mathcal{B}_s=\{s_1,..., s_k\}\in \mathcal{D}_s$.
        \item Construct binary transition matrix $\mathbf{W}$ via $\{\mathcal{N}_{s_1},... \mathcal{N}_{s_k}\}$.
        \item Train value module with $\mathcal{B}_{s}$ via $\mathcal{L}=L_s+L_{csc}$.
    \end{itemize}
    }
    
    \end{enumerate}
}
\caption{\algo{State2Emb-Ac+}}
\label{algorithm: state2emb}
\end{algorithm}

\end{document}